\documentclass{article} 
\usepackage{subcaption}
\usepackage{collas2025_conference,times}
\usepackage{easyReview}

\usepackage{amsmath,amsfonts,bm}

\def\eqref#1{equation~\ref{#1}}

\def\1{\bm{1}}

\DeclareMathAlphabet{\mathsfit}{\encodingdefault}{\sfdefault}{m}{sl}
\SetMathAlphabet{\mathsfit}{bold}{\encodingdefault}{\sfdefault}{bx}{n}

\usepackage{algorithm}
\usepackage{algorithmic}
\usepackage{booktabs} 
\usepackage{wrapfig}
\expandafter\def\csname ver@subfig.sty\endcsname{}

\usepackage{svg}
\usepackage{graphicx}
\usepackage{subcaption}
\usepackage[margin=1in]{geometry}
\usepackage{amsmath}
\usepackage{amssymb}
\usepackage{mathtools}
\usepackage{amsthm}
\usepackage{multirow}
\usepackage{titlesec} 
\usepackage{listings}
\usepackage{xcolor}

\usepackage{xcolor}
\usepackage[normalem]{ulem} 

\newif\ifshowrevisions
\showrevisionstrue

\ifshowrevisions
    
    \newcommand{\deleted}[1]{\textcolor{red}{\sout{#1}}}
\else
    
    \newcommand{\deleted}[1]{}
\fi

\definecolor{white}{rgb}{1,1,1}
\definecolor{bg}{rgb}{0.95,0.95,0.95}
\usepackage{hyperref}
\hypersetup{
    colorlinks=true,
    linkcolor=red,
    filecolor=magenta,
    urlcolor=blue,
    citecolor=purple,
    pdftitle={Overleaf Example},
    pdfpagemode=FullScreen,
    }

\title{A Simple Baseline for Stable and Plastic Neural Networks}

\collasfinalcopy 

\showrevisionstrue

\begin{document}

\author{
    Étienne Künzel\textsuperscript{\rm 1} , 
    Achref Jaziri\textsuperscript{\rm 1} , 
    Visvanathan Ramesh\textsuperscript{\rm 1,2}  \\
    \textsuperscript{\rm 1} Department of Computer Science and Mathematics, Goethe University, \\
    \textsuperscript{\rm 2} HessianAI, Frankfurt am Main, Germany \\
    \{jaziri, vramesh\}@em.uni-frankfurt.de
}

\maketitle
\begin{abstract}
Continual learning in computer vision requires that models adapt to a continuous stream of tasks without forgetting prior knowledge, yet existing approaches often tip the balance heavily toward either plasticity or stability. We introduce RDBP, a simple, low-overhead baseline that unites two complementary mechanisms: ReLUDown, a lightweight activation modification that preserves feature sensitivity while preventing neuron dormancy, and Decreasing Backpropagation, a biologically inspired gradient-scheduling scheme that progressively shields early layers from catastrophic updates. Evaluated on the Continual ImageNet benchmark, RDBP matches or exceeds the plasticity and stability of state-of-the-art methods while reducing computational cost. RDBP thus provides both a practical solution for real-world continual learning and a clear benchmark against which future continual learning strategies can be measured.
\end{abstract}
\section{Introduction}

Continual learning in computer vision tackles the fundamental challenge of enabling models to adapt to a continuous stream of visual information rather than to a single static dataset. Such systems must \emph{continuously} integrate new concepts while retaining the features and representations learned from previous tasks. This requires balancing plasticity, the ability to acquire new knowledge as data distributions shift, with stability, the capacity to preserve previously learned representations and avoid catastrophic forgetting. The plasticity–stability dilemma is especially critical in real-world applications, where access to past data may be restricted by memory constraints or privacy regulations, making rehearsal-based methods impractical \cite{mermillod2013stability,smith2023closer}.

Although many class-incremental methods report strong end-to-end performance in a continual learning setting, it often remains unclear whether those gains arise from enhanced plasticity, superior stability, or both. Recent analyses show that numerous class-incremental algorithms prioritize stability so heavily that the feature extractor trained on the initial tasks alone can be as effective as that of the full incremental model \cite{kim2023stability}. These observations  underscore the need for both a more granular understanding and analyses of continual learning dynamics and the establishment of clear, simple baselines that achieve a balanced trade-off between plasticity and stability.

A key motivation for our work is the \emph{lack of a straightforward and simple baseline} in the stability–plasticity literature. Most existing approaches skew toward optimizing either plasticity or stability in isolation. Moreover, prevalent methods for boosting plasticity such as weight perturbations that inadvertently erase previously acquired knowledge \cite{dohare2024loss, lee2024slow}, or adaptive activation functions \cite{delfosse2021adaptive} that introduce extra parameters and complex operations often complicate the incorporation of stability mechanisms and increase computational overhead.

Some recent efforts have attempted to address this by leveraging multi-scale feature encodings from pre-trained backbones, structure-wise distillation losses to preserve hierarchical representations, and bespoke stability–plasticity normalization layers to balance adaptation and retention ~\cite{jung2023new, elsayed2024addressing}. These methods have demonstrated notable performance improvements across various continual learning benchmarks, showcasing their ability to mitigate forgetting while enabling efficient task adaptation. However, despite their effectiveness, these solutions incur significant architectural complexity, rely on auxiliary distillation objectives, and depart from the simplicity desired for a baseline framework.
In this work, we explore a simple yet effective approach to continual learning that preserves high plasticity without undermining stable representations or imposing heavy computational costs. Specifically, we introduce:
\begin{itemize}
  \item \textbf{ReLUDown}, a lightweight activation modification that dynamically scales activations to enhance adaptability while leaving earlier-layer features intact;
  \item \textbf{Decreasing Backpropagation (DBP)}, a gradient-scheduling scheme inspired by biological memory consolidation processes, which progressively decreases gradient flow through lower layers to protect established knowledge.
\end{itemize}
Collectively, ReLUDown and DBP, forming our 	\textbf{RDBP} baseline, present a low-complexity framework for addressing the stability-plasticity dilemma. RDBP not only provides a practical solution for class-incremental continual learning but also serves as a clear, lightweight benchmark for comparing future continual learning approaches.

\section{Related Work}
Continual learning, also referred to as lifelong learning ~\cite{parisi2019continual}, requires models to learn from a sequence of non-stationary data streams while jointly addressing both network plasticity and stability ~\cite{kim2023stability}. Traditional neural networks, when trained sequentially on multiple tasks, often experience catastrophic forgetting ~\cite{riemer2018learning}, where previously learned representations are overwritten, and their ability to adapt effectively to new tasks is significantly reduced.

Maintaining \textbf{plasticity} ensures that models can adapt their parameters in response to a change in the data distribution \cite{berariu2021study}. Two main phenomena impact the network’s plasticity: first, shifts in the pre-activation distribution during training can cause neurons  to become “dead”   \cite{lin2015far}, “dormant” \cite{sokar2023dormant}, or effectively linearized \cite{lyle2024disentangling}, reducing the network’s expressive capacity; second, unchecked growth in parameter norms sharpens the loss landscape, driving activation and normalization layers toward saturation and diminishing sensitivity to gradient updates~\cite{damian2022self}. To counteract these effects, three main categories of methods have emerged. One line of work periodically resets or overwrites selected weights or layers to increase the network’s ability to learn new patterns~\cite{dohare2024loss,he2025plasticityawaremixtureexpertslearning}. A second approach focuses on activation-level interventions using adaptive activation functions or carefully calibrated noise injections to preserve plasticity without altering the underlying training dynamics~\cite{delfosse2021adaptive,kuo2021plastic}. Finally, architectural expansion methods introduce additional modules or subnetworks that specialize in handling new tasks, which allows the base model to retain existing representations intact~\cite{liu2025neuroplasticexpansiondeepreinforcement}.

Maintaining \textbf{stability} is essential not only to prevent catastrophic forgetting but also to facilitate efficient transfer to related tasks ~\cite{weiss2016survey,zhuang2020comprehensive}. One common strategy is to retain and replay a small set of representative examples from past tasks, thereby reinforcing old knowledge whenever the network trains on newly introduced data ~\cite{shin2017continual}. Other methods dynamically adjust the structure of the network, either by dedicating subnetworks to each task or by growing new modules on demand, to protect previously learned features ~\cite{yoon2017lifelong}. Regularization-based approaches, such as Elastic Weight Consolidation, impose penalties on updates to parameters deemed important for earlier tasks, using measures like the Fisher information matrix to identify and protect critical weights~\cite{schwarz2018progress}. 

Recent works tackle the stability-plasticity dilemma in various contexts through task sequencing, alternating update schemes, and transfer learning. For example, \citet{jaziri2024mitigating} propose multiple subspace expansions during training, which alternate between weight regularization for better stability and adaptive activations for higher plasticity in a curriculum learning setting. While \citet{scialom2022fine} employs continual fine-tuning of pre-trained models, refining large-scale features without full retraining. Integrated frameworks further combine regularization, distillation, and specialized normalization to balance adaptation and retention~\cite{jung2023new}. Yet, these methods often add architectural or computational overhead, underscoring the need for a simple baseline like RDBP proposed in this work.

\section{RDBP: Simple Baseline for the Stability-Plasticity Dilemma}
We present RDBP, that unites two complementary mechanisms to tackle the plasticity–stability dilemma. First, it employs a modified activation function, ReLUDown, designed to preserve plasticity. Second, it integrates a Decreasing Backpropagation (DBP) update scheme, which gradually decreases gradient flow through lower layers to protect previously learned weights. Together, these components form an easy-to-implement baseline that requires no auxiliary memory buffers or complex architectural additions.

\subsection{Maintaining Plasticity: ReLUDown}
\looseness=-1

For our baseline, we focus on plasticity approaches that preserve existing representations to ensure compatibility with stability mechanism. Methods that reset or expand parameters tend to impact model stability or simply defer plasticity challenges to newly added modules which can be computationally expensive in the long run. Instead, activation-based techniques maintain sensitivity to new inputs without erasing prior knowledge. Adaptive activation functions ~\cite{delfosse2021adaptive} maintain plasticity over long sequences but introduce extra computation and unstable preactivation distributions, making them hard to combine with stability mechanisms. To address these issues, we propose a static activation alternative that retains the benefits of adaptive activations.

We derive our static activation function from the standard ReLU, which fails to maintain plasticity due to the preactivation distribution gradually shifting into the negative domain, causing neurons to become dormant \cite{lyle2024disentangling}.\\
This issue can be effectively addressed by introducing a linear component with a non-zero gradient for the range from the hinge point \(d < 0 \) to \( -\infty \). This approach preserves the non-linearity of the activation function while ensuring non-zero gradients on both sides of the zero-encoding, thereby enabling the model to implicitly stabilize its preactivation distribution by the standard training process. The activation function is displayed in the appendix and can be formulated as follows:
\[f(x) =max(0,x) - max(0, -x+d) , d<0\]

\subsection{Maintaining Stability: DecreaseBackpropagation}
\begin{wrapfigure}{r}{0.5\linewidth}
    \includegraphics[width=\linewidth]{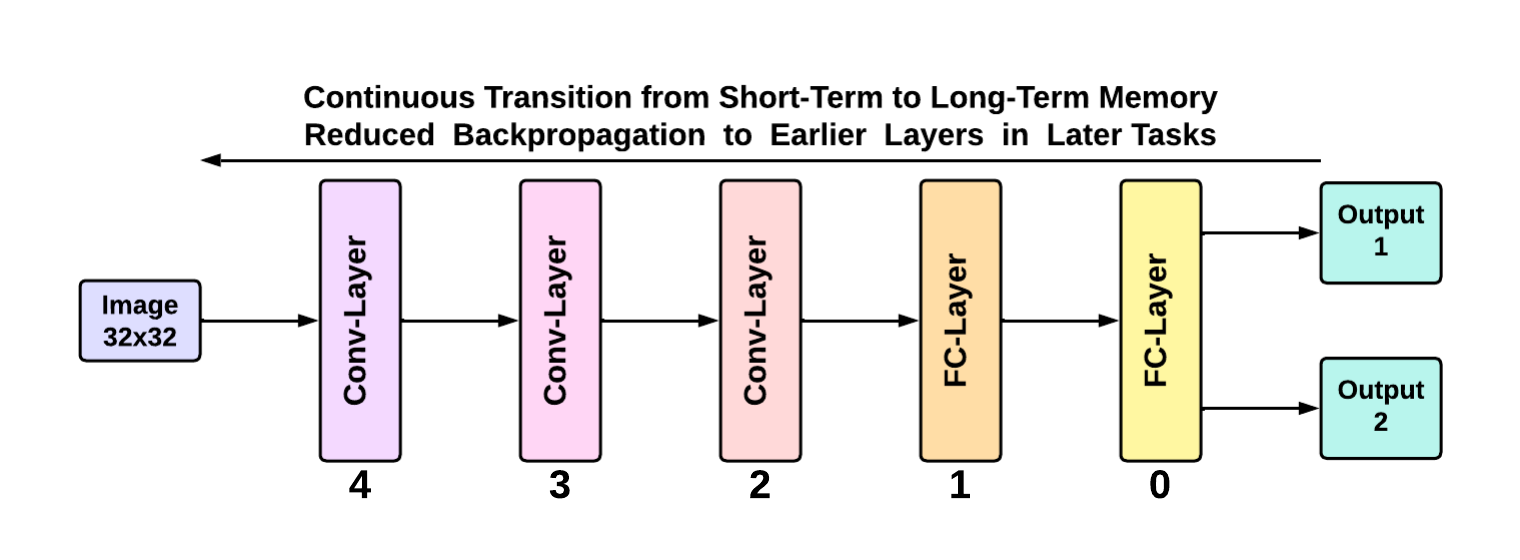}
    \caption{Schematic illustration of the DBP algorithm}
    \label{dbp}
\end{wrapfigure}
Modeling the interaction between long-term and short-term memory has been a longstanding topic of research \cite{melton1970short}. Long and short-term memory can be modeled as discrete systems \cite{izquierdo1999separate} or as points along a continuum \cite{bacsar2016brain}, each comes with its own advantages and disadvantages.\\
Our proposed solution DecreaseBackPropagation (DBP) relies on a continual modeling approach and transfers this principle to neural networks by adapting the backpropagation process. Instead of applying traditional backpropagation uniformly across all layers, our method gradually reduces the influence of backpropagation on earlier layers as the network encounters new tasks.\\
The standard backpropagation formula is adapted using an annealing process where the gradient anneals to a fraction dependent on the decrease Factor $f$ and the Layer $l$(numbered from front to back see Figure \ref{dbp}) with increasing Task Number $n$, the higher the speed factor $a$, the faster the algorithm anneals:
\[\text{bp}_{decrease}(n, l, f, a) = \text{bp}_{standard} \times ( 1-(l\times f) + (l\times f) \times a^{-n})
\]
This strategy emulates long-term memory by allowing earlier layers to encode frequently occurring patterns while maintaining adaptability in the later layers through reduced restrictions, making them flexible for the specific task.

\section{Experiments}
\subsection{Experimental Setup and Metrics}
\looseness=-1
 We evaluate our models on  an adapted version of the ImageNet Dataset \cite{deng2009imagenet} called Continual Imagenet \cite{dohare2024loss}, where each task is  binary classification   between two ImageNet classes. Each model has only has access to the data of the current task. We contrast the proposed method with  Continual Backpropagation, which prioritizes network plasticity \cite{dohare2024loss} and Generative Replay, which prioritizes preserving prior task knowledge\cite{shin2017continual}. For RDBP, we chose a hinge point of $d = -3$, a decrease factor of $f = 0.15$ and a speed factor of $a=1.005$. All networks use 3 convolution layers followed by two fully connected layers. For further information about the baselines, hyperparameters and evaluation variables see the Appendix. \\
We assess plasticity by measuring the accuracy of the model in each newly learned task immediately after training in a continual sequence. The classification head is reset after each task and stored. Stability is quantified as the mean accuracy over the previous ten tasks, evaluated without any further weight updates to the network and the classification head of the matching task.

\subsection{Results and Discussion}
\looseness=-1
\begin{figure}[ht]
\vspace{-3.5mm}
  \centering
\begin{minipage}[t]{0.51\textwidth}
  \vspace{0pt} 
  \includegraphics[width=\textwidth]{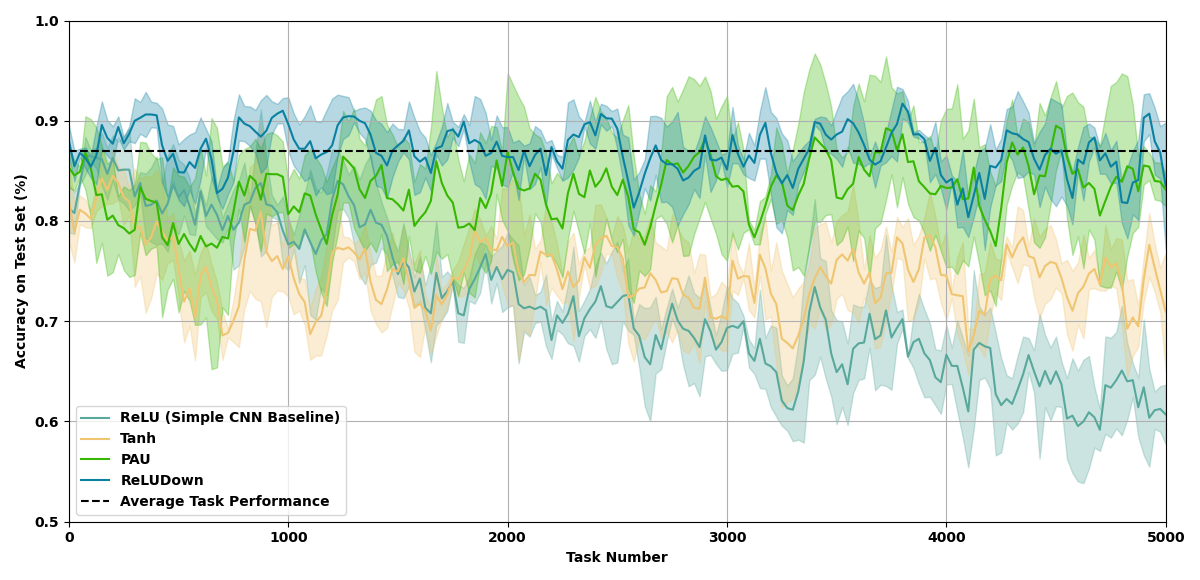}
\end{minipage}
\hfill
\begin{minipage}[t]{0.48\textwidth}
  \vspace{0pt} 
    \begin{minipage}[t]{\textwidth}
      \centering
      \begin{subfigure}[t]{0.24\textwidth}
        \includegraphics[width=\linewidth]{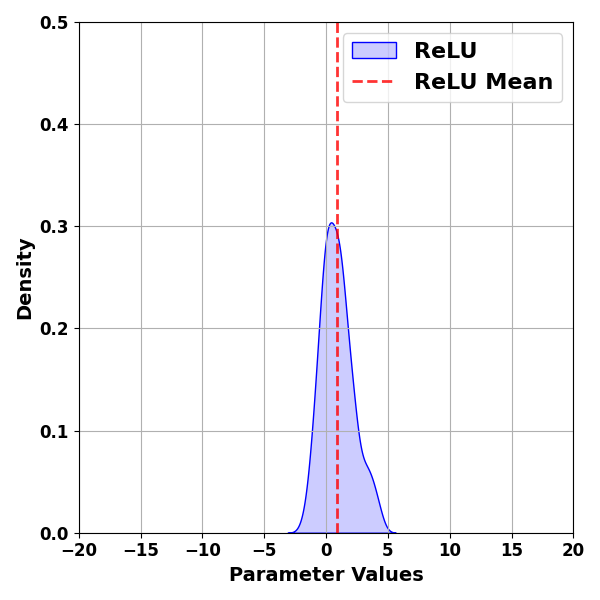}
      \end{subfigure}
      \hfill
     \begin{subfigure}[t]{0.24\textwidth}
        \includegraphics[width=\linewidth]{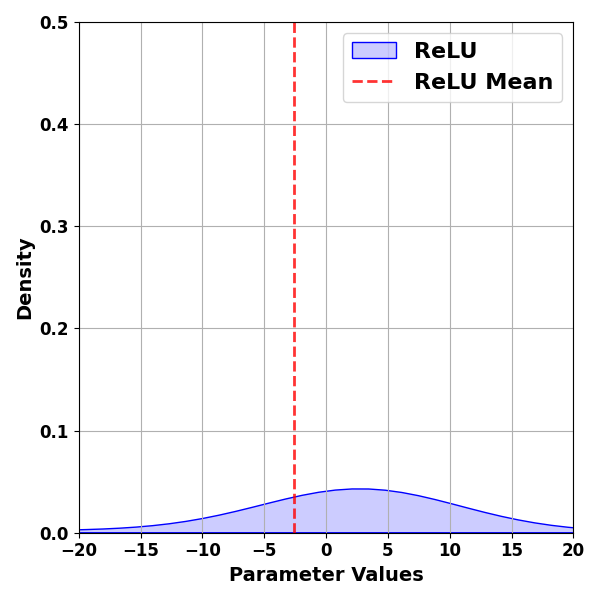}
      \end{subfigure}
      \hfill
      \begin{subfigure}[t]{0.24\textwidth}
        \includegraphics[width=\linewidth]{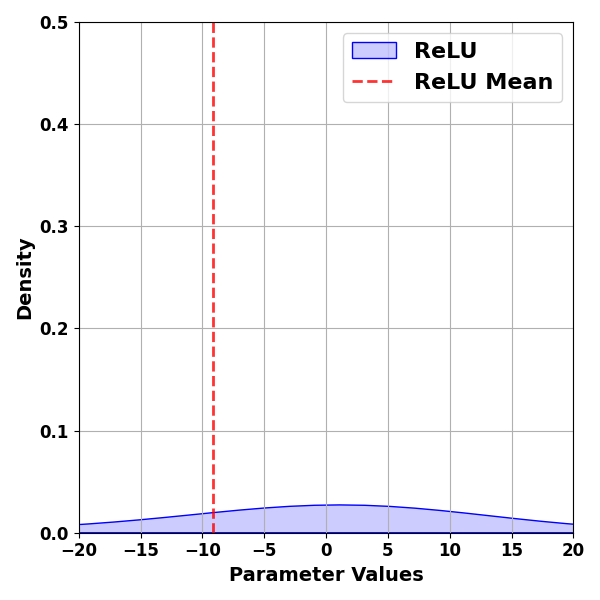}
      \end{subfigure}
      \hfill
      \begin{subfigure}[t]{0.24\textwidth}
        \includegraphics[width=\linewidth]{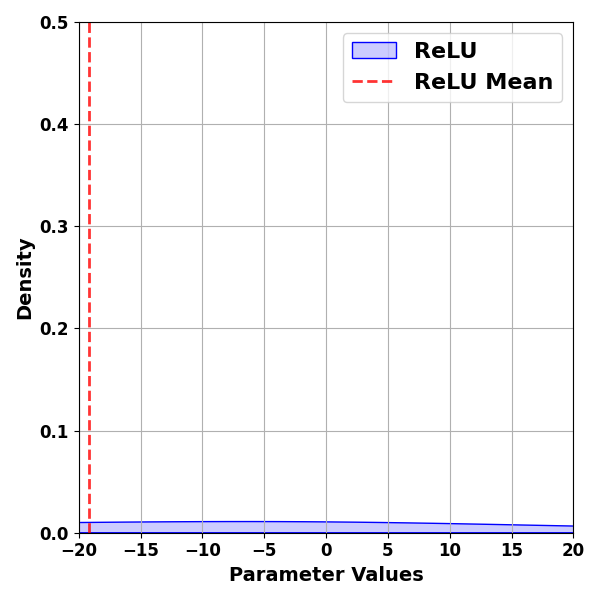}
      \end{subfigure}
    \end{minipage}

    \begin{minipage}[t]{\textwidth}
      \centering
      \begin{subfigure}[t]{0.24\textwidth}
        \includegraphics[width=\linewidth]{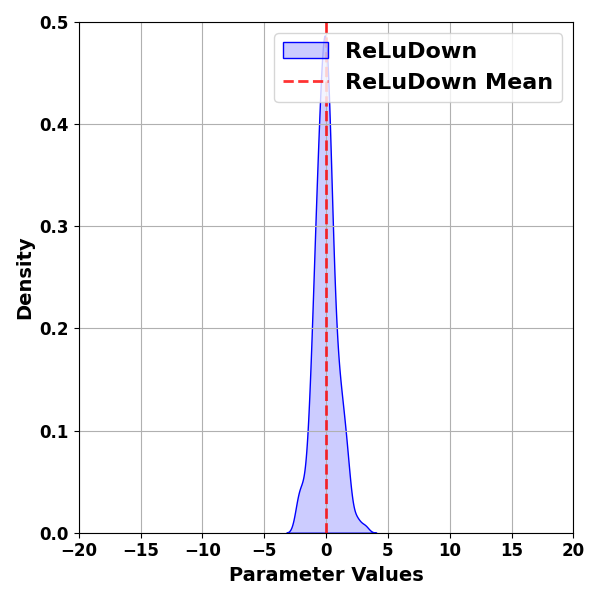}
      \end{subfigure}
      \hfill
     \begin{subfigure}[t]{0.24\textwidth}
        \includegraphics[width=\linewidth]{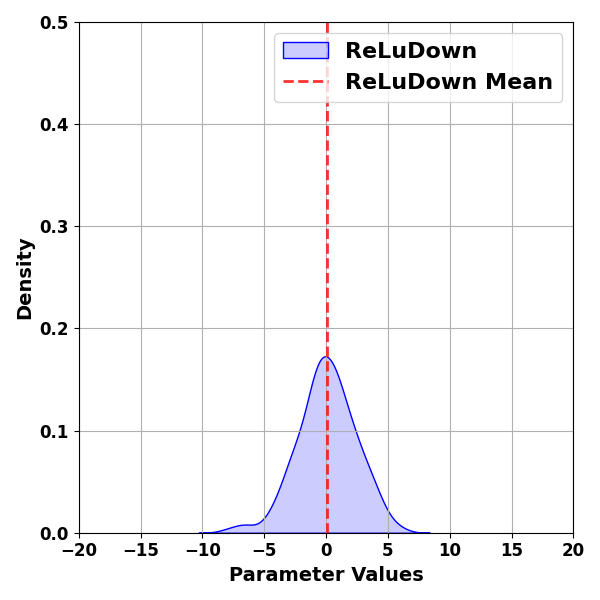}
      \end{subfigure}
      \hfill
      \begin{subfigure}[t]{0.24\textwidth}
        \includegraphics[width=\linewidth]{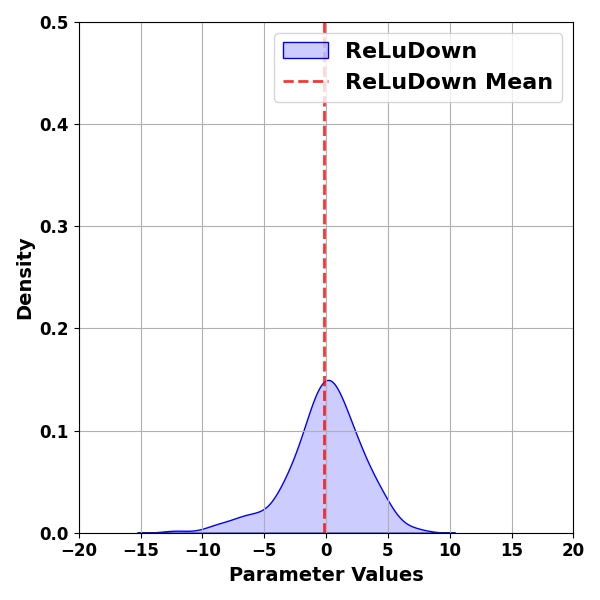}
      \end{subfigure}
      \hfill
      \begin{subfigure}[t]{0.24\textwidth}
        \includegraphics[width=\linewidth]{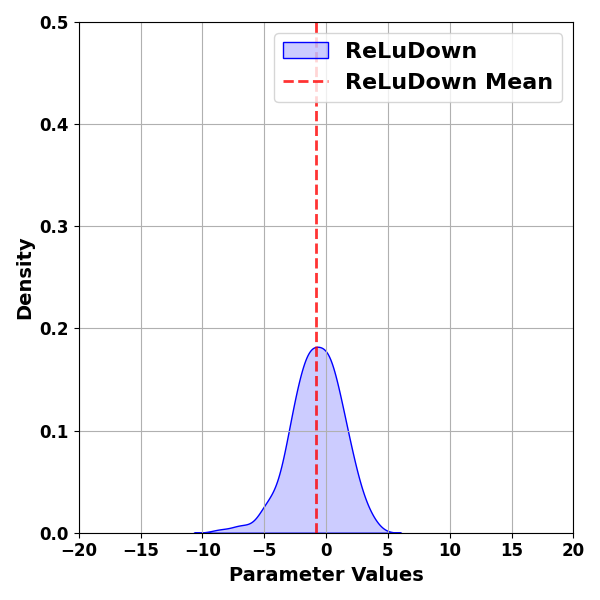}
      \end{subfigure}
    \end{minipage}
  \end{minipage}

  \caption{\textbf{Left}: Current Task Performance of a CNN with different Activation Functions: ReLU, Tanh, ReLUDown,  PAU and the Average Performance of a CNN fully reset at each task \textbf{Right}: Preactivation Distribution for Task 100, 1000, 3000, 5000 of the CNN with the ReLU(top) and ReLUDown(bottom)}
  \label{plasact}
\end{figure}
In Figure \ref{plasact}, we illustrate the impact of various activation functions on the neural network’s ability to maintain plasticity (left panel). The results demonstrate that the network utilizing the ReLU activation function exhibits a decline in performance, despite similar task complexity. In contrast, networks employing the ReLUDown or Padé Activation Unit\cite{delfosse2021adaptive} activation functions are able to maintain their plasticity over time. Additionally, our proposed activation function achieves this while also reducing training time by approximately 20\% compared to the Padé Activation Units (PAU)(see Appendix \ref{time}).\\
The right panel of Figure \ref{plasact} provides insights into the preactivation distribution. In the top row, we can see the network with the ReLU activation function experiences a shift in the preactivation distribution shift. Conversely, for the ReLUDown, the preactivation distribution converges toward a normal distribution, thereby mitigating one of the main mechanisms of plasticity loss: "Linearization and preactivation distribution shift" noted by \citet{lyle2024disentangling}.

\begin{figure}[ht]
  \centering
\begin{minipage}[t]{0.49\textwidth}
  \vspace{0pt} 
  \includegraphics[width=\textwidth]{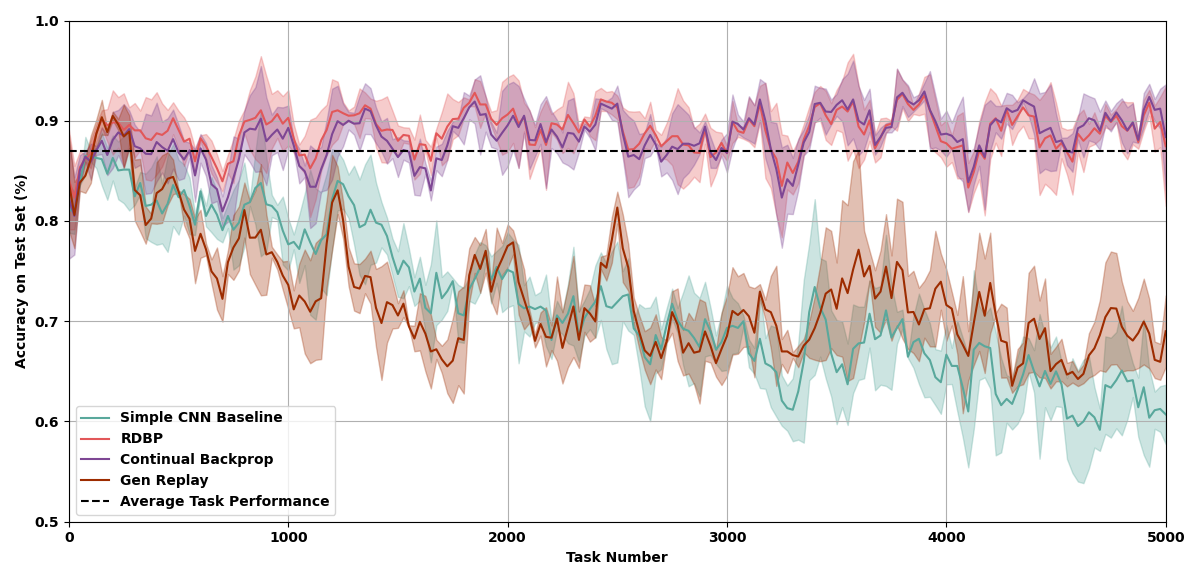}
\end{minipage}
\hfill
\begin{minipage}[t]{0.49\textwidth}
  \vspace{0pt} 
  \includegraphics[width=\textwidth]{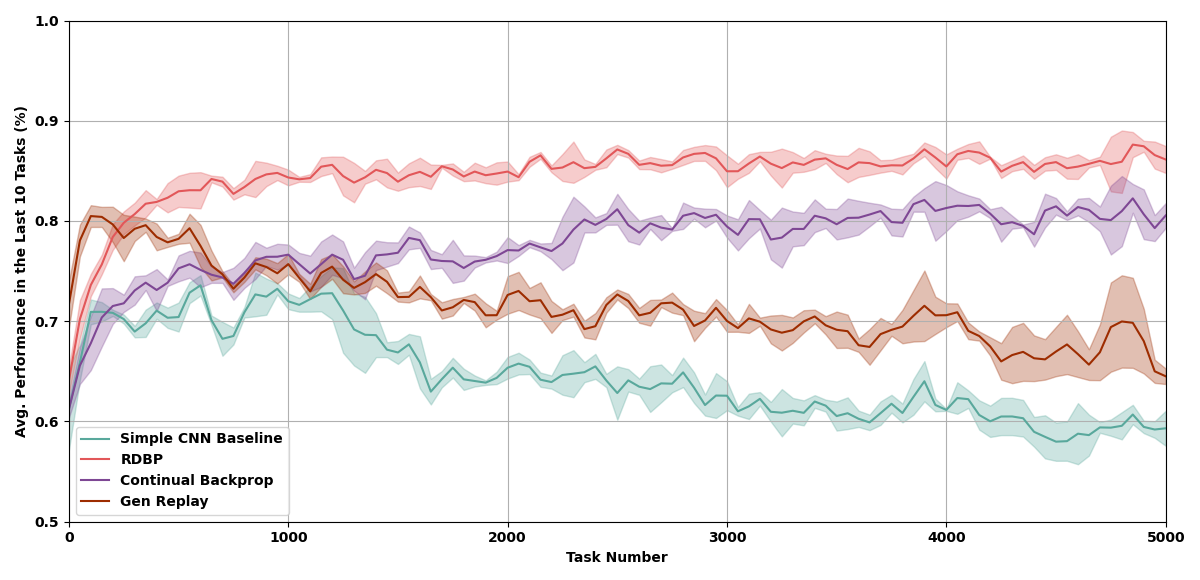}
\end{minipage}
  \caption{Performance of a CNN with a ReLU, RDBP, Continual Backpropagation and Generative Replay.
  \textbf{Left}:Current Task Performance to evaluate Plasticity.
  \textbf{Right}: Average Performance in the last 10 Tasks to evaluate Stability }
  \label{stabo}
\end{figure}
In Figure \ref{stabo}, we compare our approach RDBP with Standard CNN, Continual Backpropagation, and Generative Replay in terms of network plasticity (left) and stability(right).\\
For plasticity, RDBP performs similarly to Continual Backpropagation with both of their performances fluctuating around the performance level network trained on each task independently oscillate around the single‐task training benchmark (black dotted line), demonstrating preserved plasticity and effective transfer learning in some cases. While Continual Backpropagation achieves this with neuron resets our approach achieves similar results without additional computations. When using Generative Replay with a ReLU activation function, the classifier still loses plasticity like the Simple CNN baseline, though slightly more slowly, likely due to increased variance in the training data, which makes it harder for the classifier to overfit to individual tasks. \\
For stability, RDBP showcases an increasing performance with progressing training due to the fact that in earlier tasks the gradients are not as reduced as in later tasks. Continual Backpropagation maintains plasticity by resetting parameters, yet remains stable, likely due to a high maturity threshold preserving knowledge of the last ten tasks, but suffers catastrophic forgetting for older tasks. Generative Replay shows good task maintenance for the first tasks, which then decreases due to the decline in plasticity, underlining the importance of considering both plasticity and stability when evaluating continual learning methods.
\section{Conclusion and Future Work}

We introduce ReLUDown, a drop-in activation that increases plasticity in continuous learning by preserving the dynamics of the standard network, avoiding pre-activation changes and requiring no parameter changes.  Paired with Decreasing Backpropagation (DBP), which gradually attenuates gradient updates in early layers to mirror a continuum between short- and long-term memory and enhance stability, these two components form RDBP.  RDBP balances plasticity and stability in an incremental task setting, all without extra memory, architectural tweaks, or added complexity. We hypothesize that this algorithm will perform effectively in settings where the underlying task structure remains consistent. This is particularly relevant in domains where representations obtained from convolutional layers are transferable across tasks (i.e., object classification). For future work, we will systematically benchmark RDBP and various continual learning methods across a wider spectrum of incremental paradigms, including classes, tasks, and domain incremental settings. Moreover, for applications where gradient attenuation alone cannot fully prevent forgetting, we will investigate hybrid schemes that couple RDBP with stronger stability modules, such as generative replay buffers or adaptive network expansion, to dynamically reinforce memory retention without sacrificing plasticity.
\section*{Acknowledgments}
This work was supported by the KIBA Project (Künstliche Intelligenz und diskrete Beladeoptimierungsmodelle zur Auslastungssteigerung im Kombinierten Verkehr - KIBA) under reference number 45KI16E051, funded by the German Ministry of Transportation.



\bibliography{collas2025_conference}
\bibliographystyle{collas2025_conference}

\appendix
\section{Appendix}

\subsection{Evaluation Setting}

\begin{figure}[h]
\vspace{-5mm}
    \centering
    \includegraphics[width=1\linewidth]{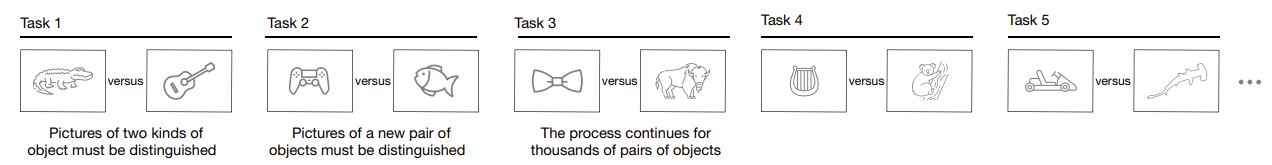}
    \caption{Continual ImageNet \cite{dohare2024loss}}
    \label{fig:enter-label}
    \vspace{-2mm}

\end{figure}
The Continual ImageNet benchmark uses a downsampled 32×32 version of ImageNet, consisting of 1,000 classes with 700 images per class(600 for training and 100 for testing). Each binary classification task involves training on 1,200 images and testing on 200 images. Models are trained over 250 epochs using mini-batches of 100. We trained for 5 runs of 5000 tasks each.
\subsection{Example Images}
\begin{figure}[ht]
\vspace{-5mm}
  \centering
\begin{minipage}[t]{0.19\textwidth}
  \vspace{0pt} 
  \includegraphics[width=\textwidth]{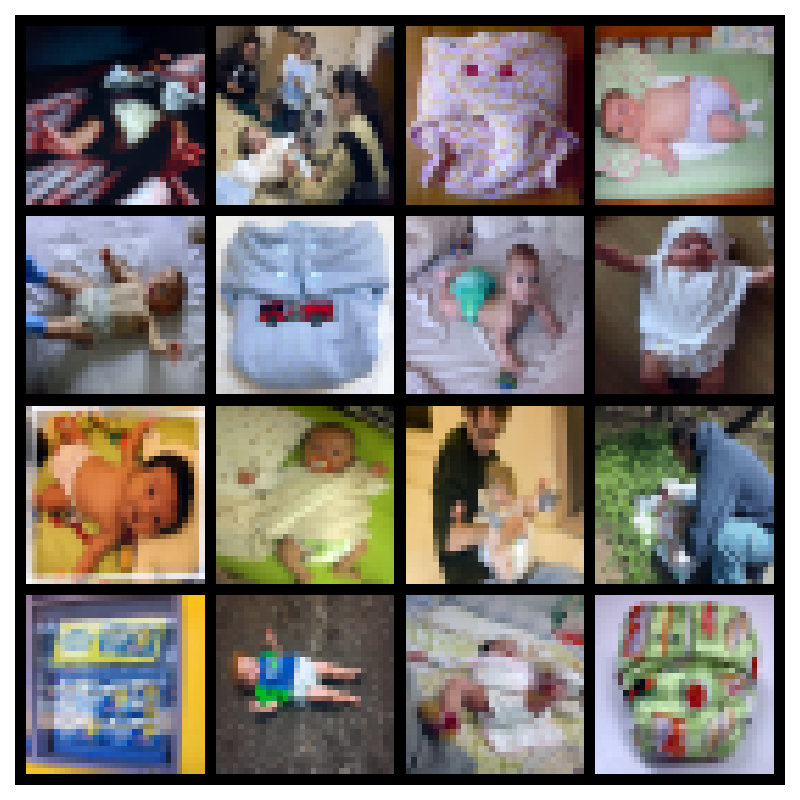}
\end{minipage}
\hfill
\begin{minipage}[t]{0.19\textwidth}
  \vspace{0pt} 
  \includegraphics[width=\textwidth]{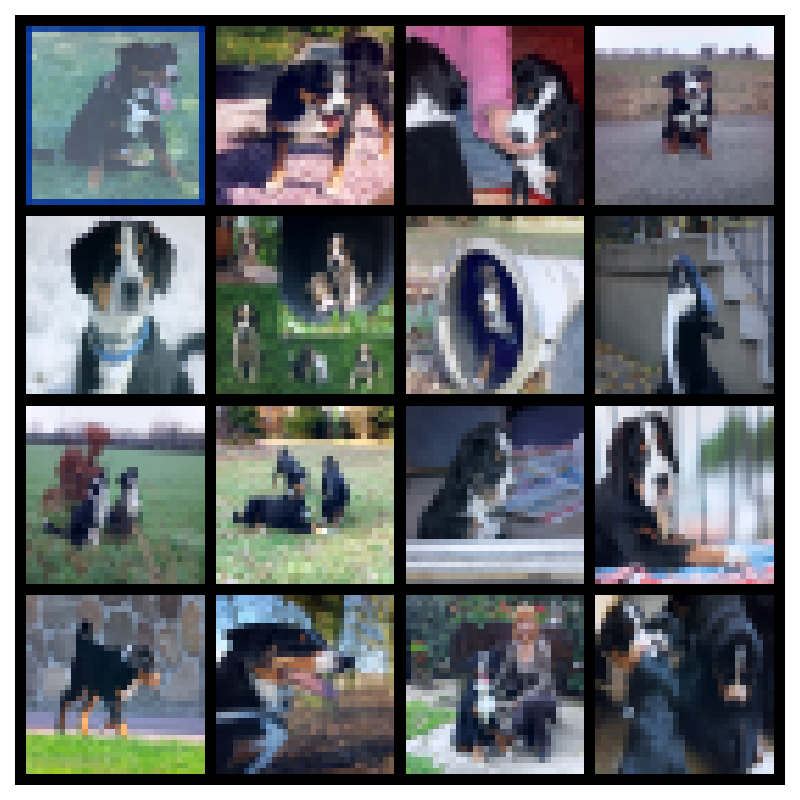}
\end{minipage}
\begin{minipage}[t]{0.19\textwidth}
  \vspace{0pt} 
  \includegraphics[width=\textwidth]{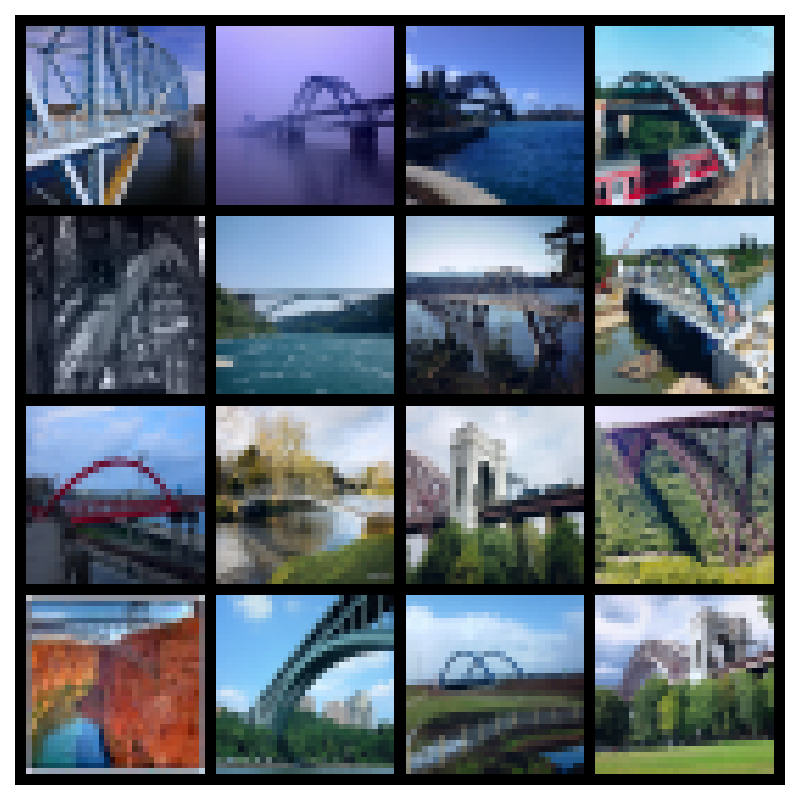}
\end{minipage}
\begin{minipage}[t]{0.19\textwidth}
  \vspace{0pt} 
  \includegraphics[width=\textwidth]{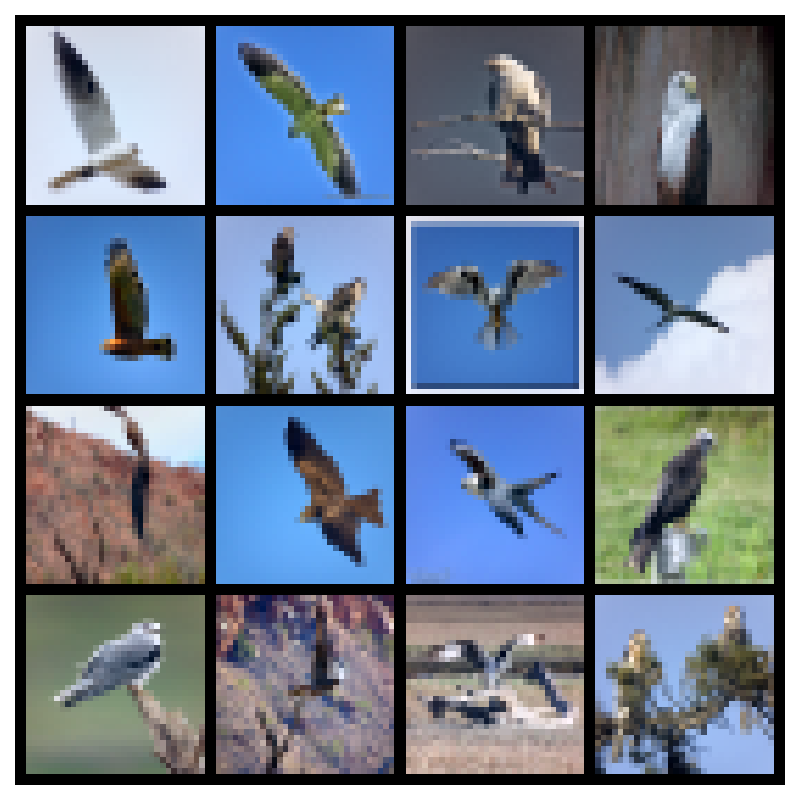}
\end{minipage}
\begin{minipage}[t]{0.19\textwidth}
  \vspace{0pt} 
  \includegraphics[width=\textwidth]{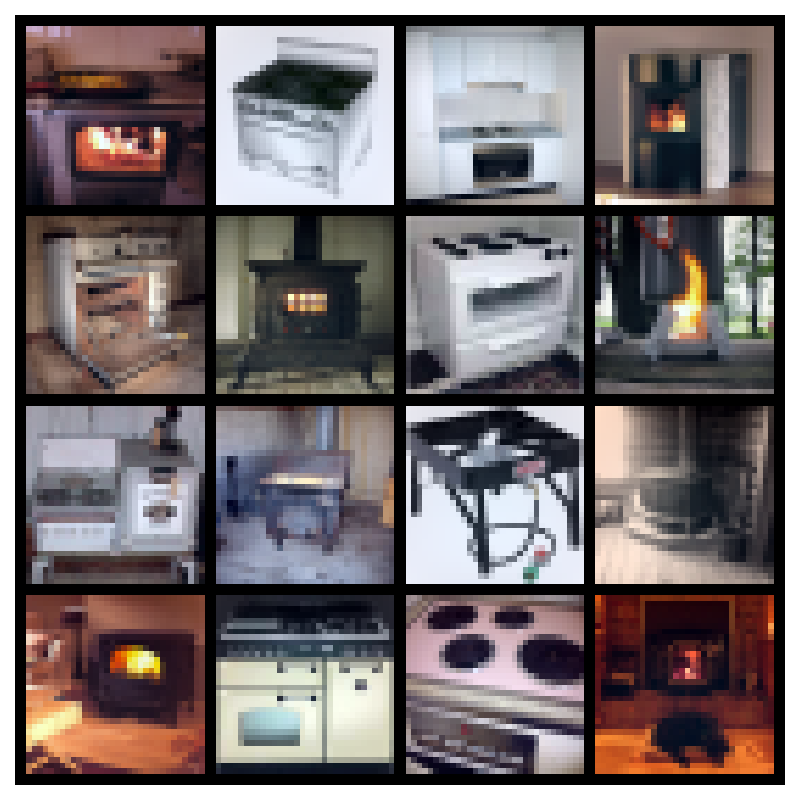}
\end{minipage}
  \caption{Example Images of the downsampled ImageNet Dataset. : Baby, Dog, Bridge, Bird, Oven}
\end{figure}

\subsection{Activation Functions}
\begin{figure}[ht]
  \centering
\begin{minipage}[t]{0.24\textwidth}
  \vspace{0pt} 
  \includegraphics[width=\textwidth]{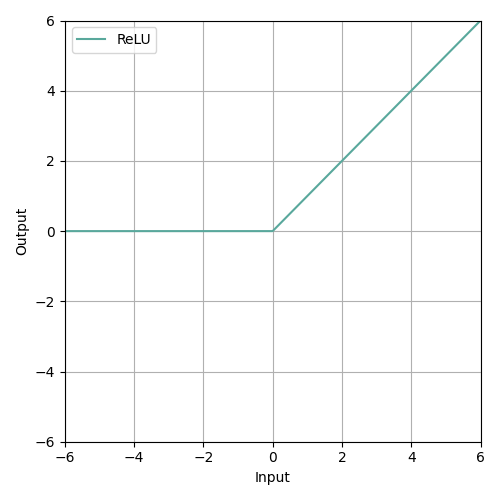}
\end{minipage}
\hfill
\begin{minipage}[t]{0.24\textwidth}
  \vspace{0pt} 
  \includegraphics[width=\textwidth]{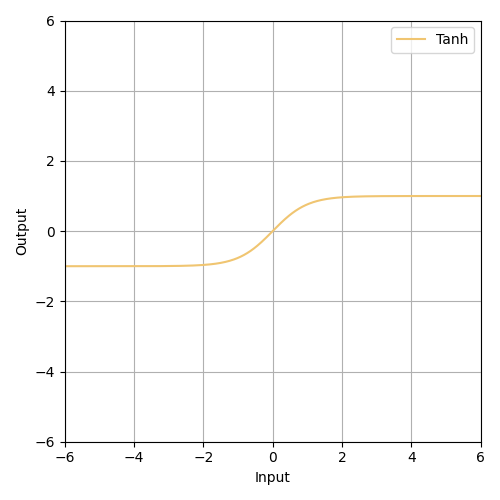}
\end{minipage}
\begin{minipage}[t]{0.24\textwidth}
  \vspace{0pt} 
  \includegraphics[width=\textwidth]{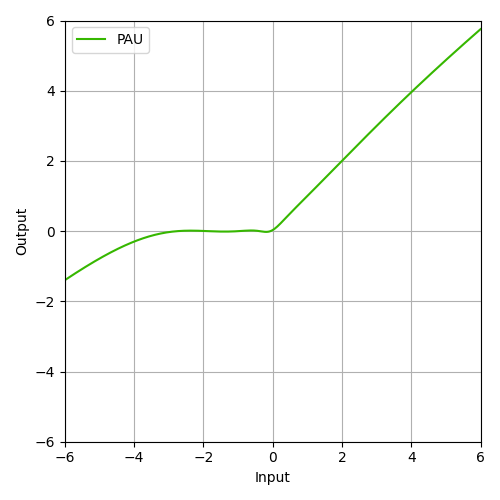}
\end{minipage}
\begin{minipage}[t]{0.24\textwidth}
  \vspace{0pt} 
  \includegraphics[width=\textwidth]{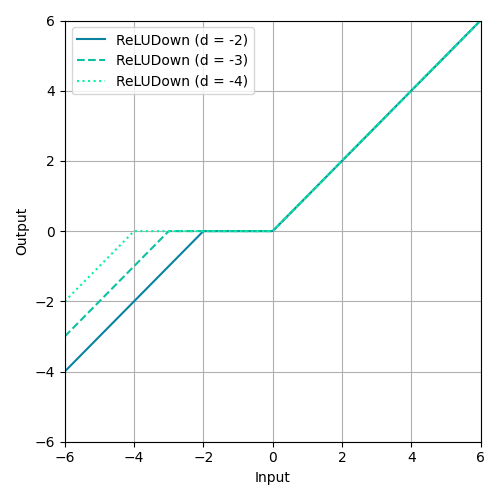}
\end{minipage}
  \caption{Activation Function used in the Experiments: ReLU, Tanh, PAU, ReLUDown}
  \label{func}
\end{figure}

Figure \ref{func} illustrates the activation functions used in our experiments. The first is the standard Rectified Linear Unit, followed by the hyperbolic tangent function. The third is the initialization of the Pade Activation Unit, which we chose to initialize as a ReLU, but then adapts during training.\\
Lastly, our proposed activation function, ReLUDown, is shown with three distinct hinge points from which we only used d=-3.

\subsection{Decrease Backpropagation}
\begin{wrapfigure}{r}{0.5\linewidth}
    \vspace{-5mm}
    \includegraphics[width=\linewidth]{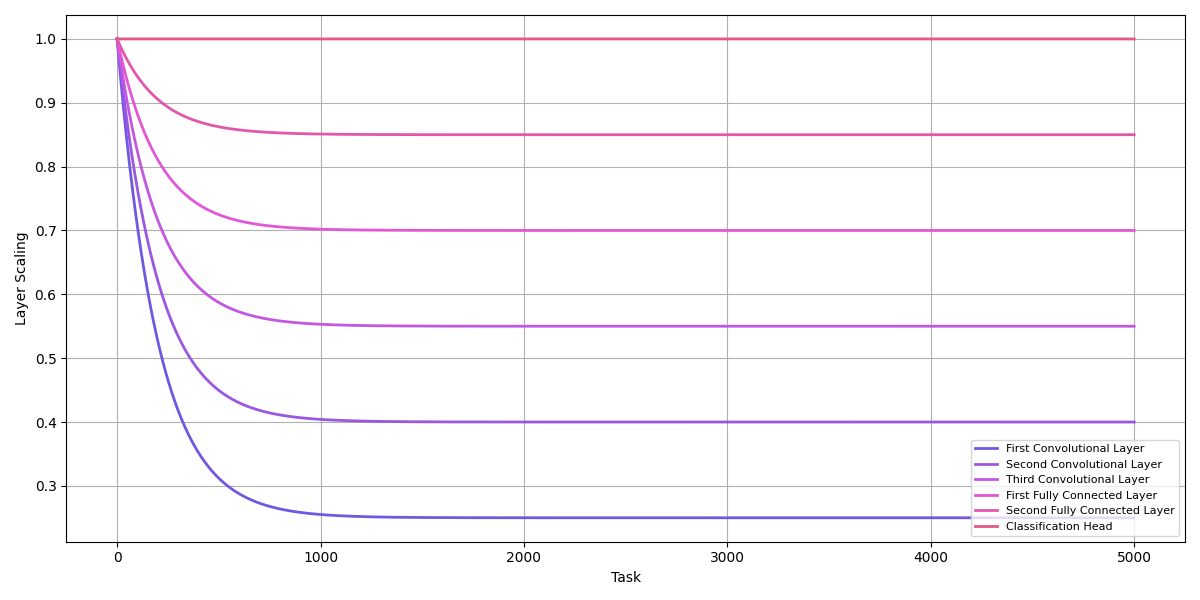}
    \caption{Gradient Factor for the different Layers of the DBP algorithm with a Speed Factor of $a = 1.005$ and a Decrease Factor of $f = 0.15$ based on the task.}
    \label{dddbp}
\end{wrapfigure}
In Figure \ref{dddbp}, we observe a decline in backpropagation activity for our algorithm (RDBP). All layers start with the same gradient factor of one, making it able to learn the basic structures of the environment. With further training, the gradient factor is reduced. After Task 1000, the gradient factor stabilizes, with the first convolutional layer receiving only 25\% of the backpropagation signal compared to a standard network. In contrast, the classification head maintains full adaptability with no reduction in gradient flow.\\
It will be interesting to investigate how altering the hyperparameters for the Speed Factor and Decrease Factor affects the algorithm’s performance in retaining and acquiring knowledge, as well as to explore the potential impact of alternative annealing strategies.

\newpage
\subsection{Results}
\subsubsection{Plasticity}
\begin{figure}[ht]
  \centering
\begin{minipage}[t]{0.49\textwidth}
  \vspace{0pt} 
  \includegraphics[width=\textwidth]{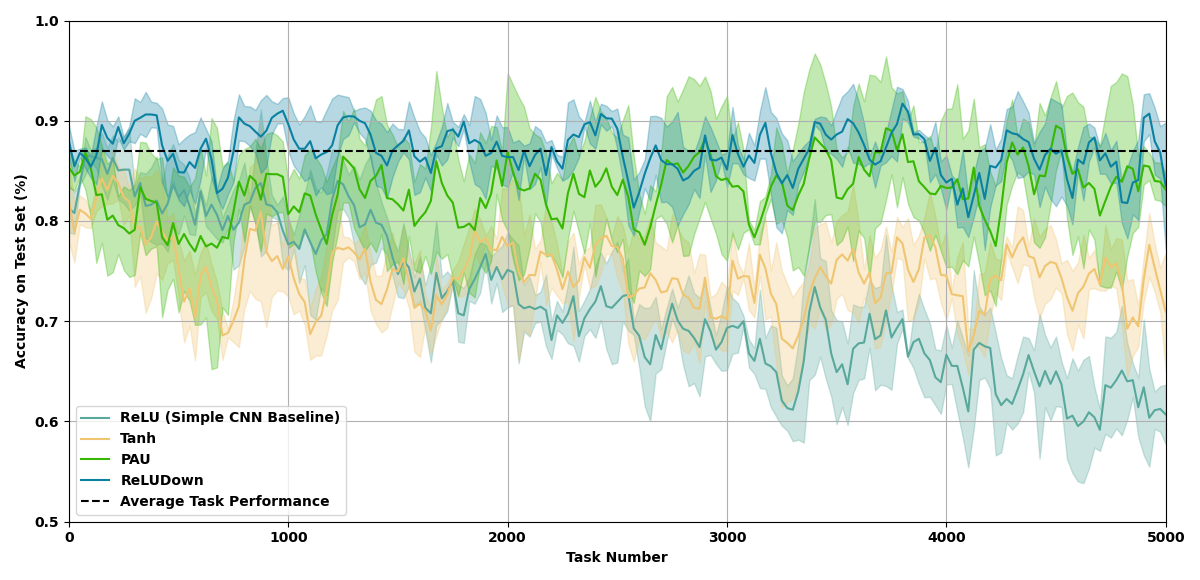}
\end{minipage}
\hfill
\begin{minipage}[t]{0.49\textwidth}
  \vspace{0pt} 
  \includegraphics[width=\textwidth]{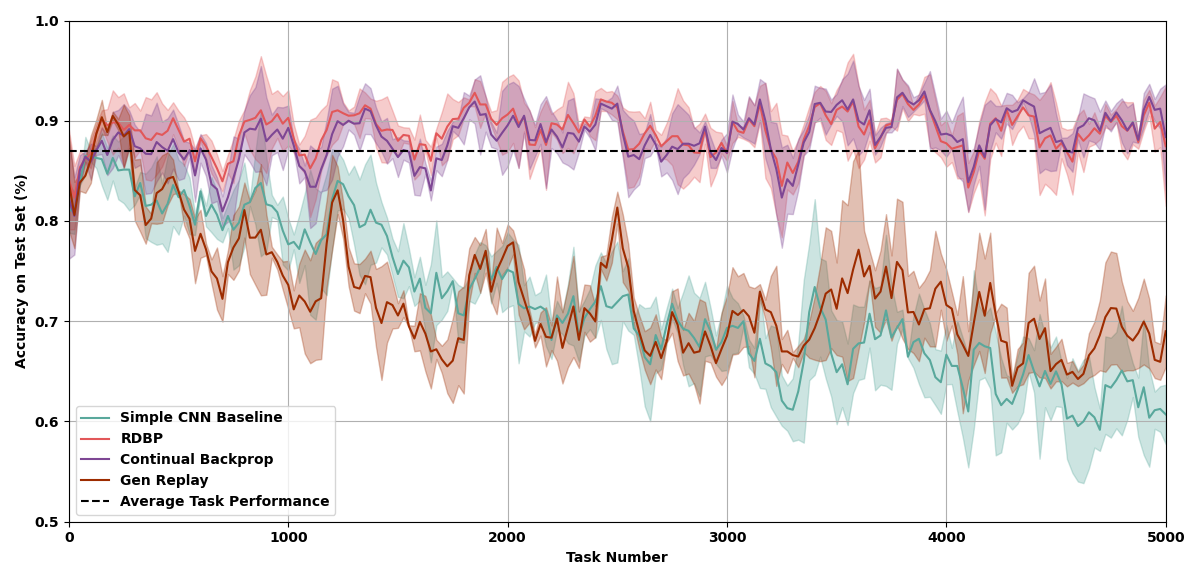}
\end{minipage}
  \caption{Current Task Performance of a CNN to evaluate plasticity with different Activation Functions: ReLU, Tanh, ReLUDown,  PAU on the left and RDBP, Continual Backpropagation, and Generative Replay on the right. The average Performance of a CNN fully reset at each task is included as a black dotted line.}
\end{figure}

\subsubsection{Stability}
\begin{figure}[ht]
  \centering
\begin{minipage}[t]{0.49\textwidth}
  \vspace{0pt} 
  \includegraphics[width=\textwidth]{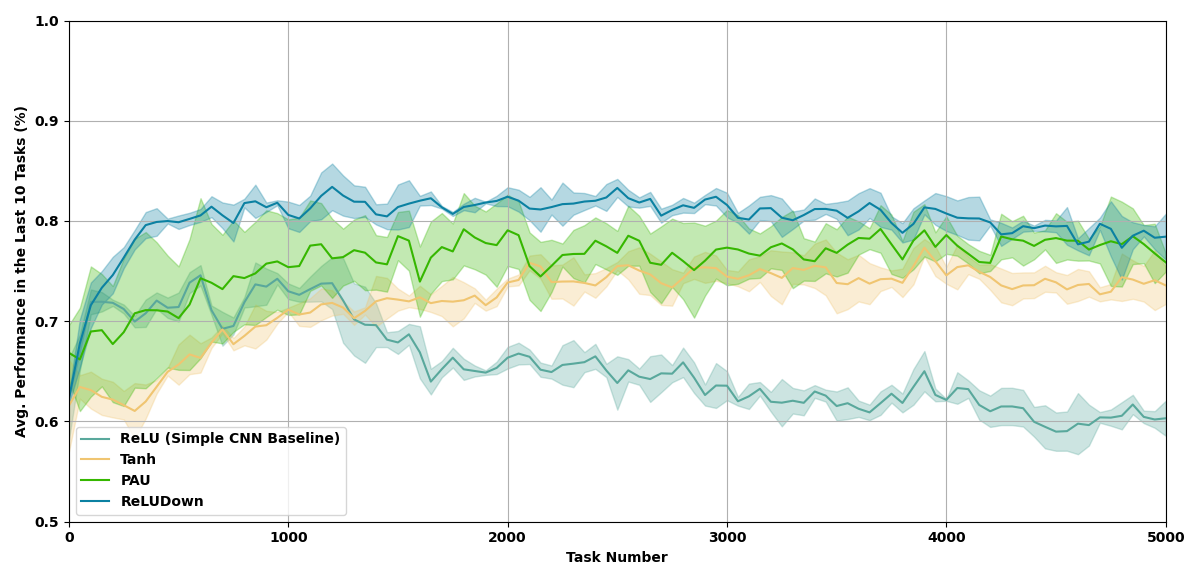}
\end{minipage}
\hfill
\begin{minipage}[t]{0.49\textwidth}
  \vspace{0pt} 
  \includegraphics[width=\textwidth]{figures/StabilityFigures/stability2.png}
\end{minipage}
  \caption{ Average Performance in the last 10 Tasks to evaluate the Stability of a CNN with different Activation Functions: ReLU, Tanh, ReLUDown,  PAU on the left and RDBP, Continual Backpropagation, and Generative Replay on the right.}
  \label{stabo2}
\end{figure}

\subsubsection{Time}
\label{time}
\begin{wrapfigure}{r}{0.5\textwidth}
\centering
\vspace{-1em}  
\begin{tabular}{|l|l|c|}
\hline
Algorithm & Function & Rel. 
Train Time\\
\hline
Conv-Net & ReLU & 0\% \\
Conv-Net & Tanh & 1\% \\
Conv-Net & PAU & 60\%\\
Conv-Net & ReLUDown & 34\% \\
\hline
Decrease Backprop. & ReLUDown & 35\% \\
Continual Backprop. & ReLU & 26\% \\
Generative Replay& ReLU & 81\%\\
\hline
\end{tabular}
\caption{Training time per task for different algorithms.}
\label{tab:training_time}
\vspace{-5mm}
\end{wrapfigure}
Table \ref{tab:training_time} presents the relative training times of various activation functions and continual learning strategies, expressed as percentage increases compared to a baseline convolutional network using ReLU.\\
The Tanh activation function introduces a negligible computational overhead. In contrast, PAUs result in a substantial 60\% increase in training time, attributed to the additional trainable parameters that are continuously updated during learning.\\
The ReLUDown activation function leads to a 34\% increase in training time relative to standard ReLU. We hypothesize that the relatively low training time of the convolutional network with ReLU is due to the predominantly zero gradients of dormant neurons in later tasks, a problem our activation function does not suffer.\\
Applying Decrease Backpropagation in conjunction with ReLUDown does not result in any further significant overhead beyond that introduced by ReLUDown alone.\\
Finally, Generative Replay using a VAE incurs the highest computational cost, increasing training time by 81\% compared to standard ReLU. This is due to the necessity of training a generative model alongside the classifier.

\newpage
\subsubsection{Preactivation Distributions of Activation Functions}
In this section, we visualize and compare the preactivation distributions observed across ReLU, Tanh, PAU, and ReLUDown(Ours). This analysis provides insights into how each function responds to a shifting input distribution. We took representative distributions for Task 0, 100, 500, 1000, 3000 and 5000.\\
When using the ReLU activation function, the preactivation distribution exhibits a consistent shift toward negative values and becomes increasingly dispersed. As a result, most activations are zero with zero gradients. This worsens adaptation to new tasks and learning performance.\\
Although Tanh performs worse than ReLU on the initial tasks, it retains more plasticity over time, allowing it to outperform ReLU in later tasks. This is reflected in its preactivation distribution, which stays at a mean of zero with a growing standard deviation.\\
The preactivation distribution of the Network with PAUs stays around a zero mean but displays fluctuations of the distribution shape.\\
ReLUDown begins with a similar preactivation distribution as other activation functions. However, unlike them, its distribution gradually converges toward a near-normal shape, exhibiting only minor fluctuations in both mean and standard deviation.
\subsubsection{Relu}
\begin{figure}[H]
\begin{minipage}[t]{01\textwidth}
  \vspace{0pt} 
    \begin{minipage}[t]{\textwidth}
      \centering
      \begin{subfigure}[t]{0.16\textwidth}
        \includegraphics[width=\linewidth]{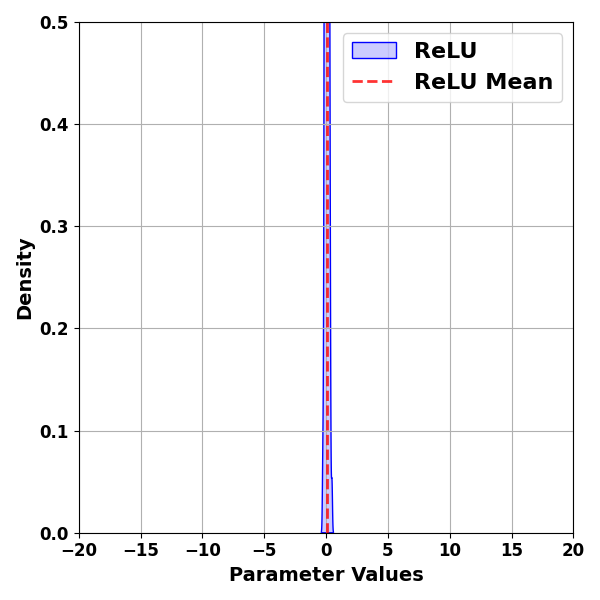}
      \end{subfigure}
      \hfill
     \begin{subfigure}[t]{0.16\textwidth}
        \includegraphics[width=\linewidth]{figures/activations/relu/frame_100.png}
      \end{subfigure}
      \hfill
      \begin{subfigure}[t]{0.16\textwidth}
        \includegraphics[width=\linewidth]{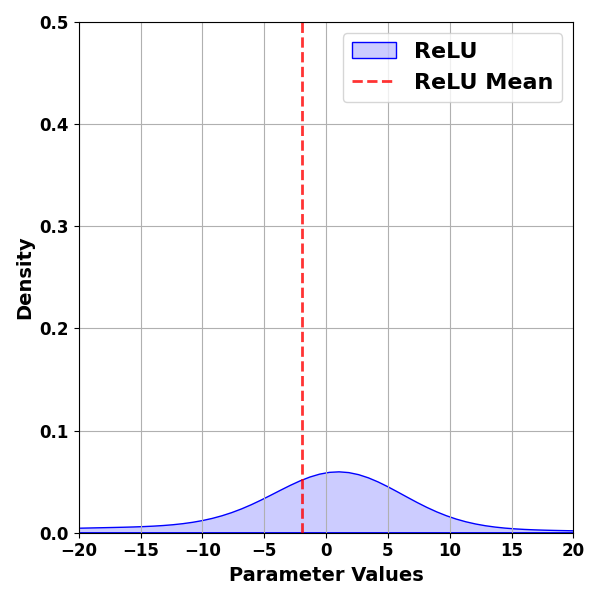}
      \end{subfigure}
      \hfill
      \begin{subfigure}[t]{0.16\textwidth}
        \includegraphics[width=\linewidth]{figures/activations/relu/frame_1000.png}
      \end{subfigure}
      \hfill
      \begin{subfigure}[t]{0.16\textwidth}
        \includegraphics[width=\linewidth]{figures/activations/relu/frame_3000.png}
      \end{subfigure}
      \hfill
      \begin{subfigure}[t]{0.16\textwidth}
        \includegraphics[width=\linewidth]{figures/activations/relu/frame_5000.png}
      \end{subfigure}
    \end{minipage}
  \end{minipage}
\end{figure}

\subsubsection{Tanh}
\begin{figure}[H]
\begin{minipage}[t]{01\textwidth}
  \vspace{0pt} 
    \begin{minipage}[t]{\textwidth}
      \centering
      \begin{subfigure}[t]{0.16\textwidth}
        \includegraphics[width=\linewidth]{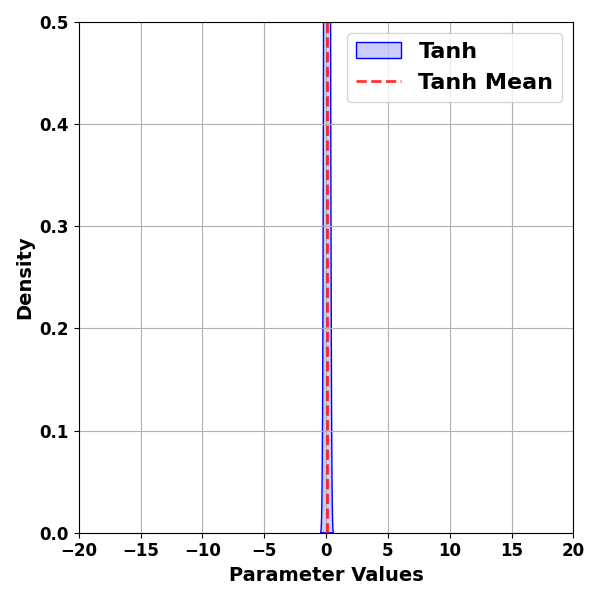}
      \end{subfigure}
      \hfill
     \begin{subfigure}[t]{0.16\textwidth}
        \includegraphics[width=\linewidth]{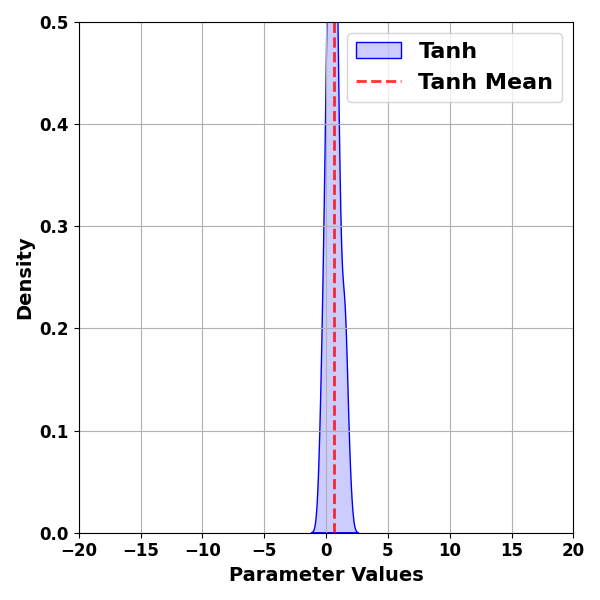}
      \end{subfigure}
      \hfill
      \begin{subfigure}[t]{0.16\textwidth}
        \includegraphics[width=\linewidth]{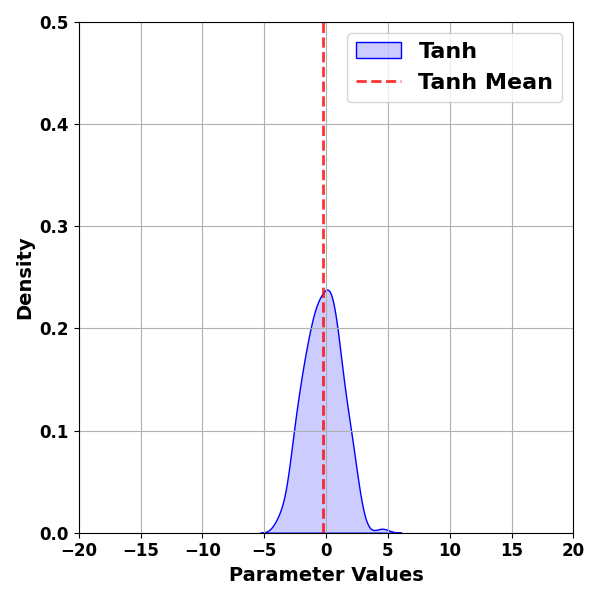}
      \end{subfigure}
      \hfill
      \begin{subfigure}[t]{0.16\textwidth}
        \includegraphics[width=\linewidth]{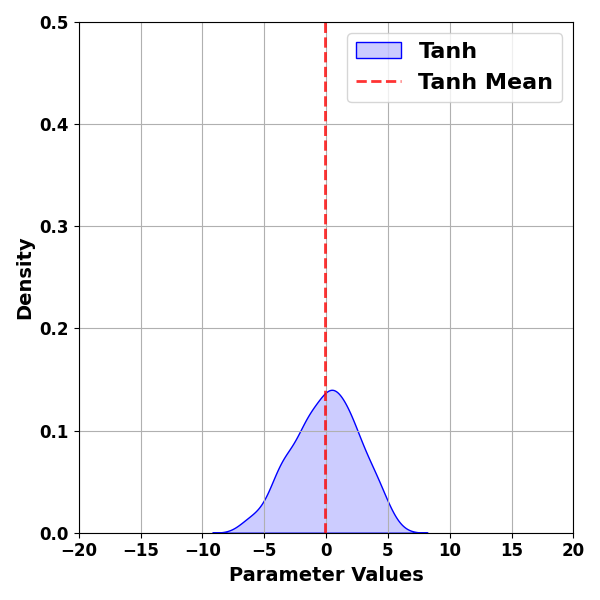}
      \end{subfigure}
      \hfill
      \begin{subfigure}[t]{0.16\textwidth}
        \includegraphics[width=\linewidth]{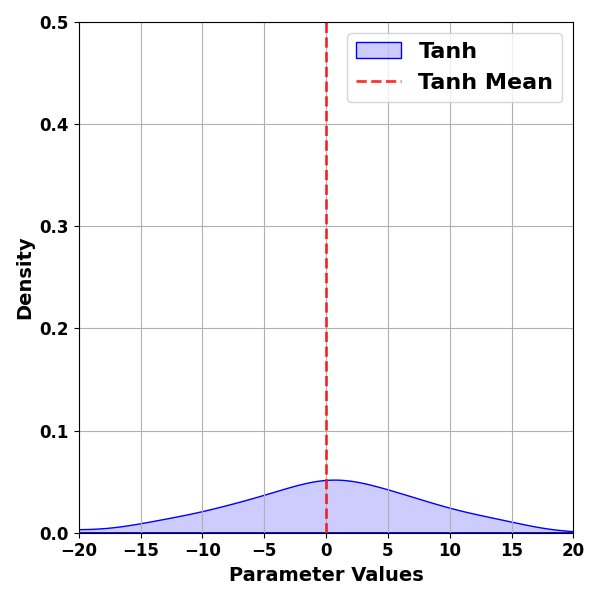}
      \end{subfigure}
      \hfill
      \begin{subfigure}[t]{0.16\textwidth}
        \includegraphics[width=\linewidth]{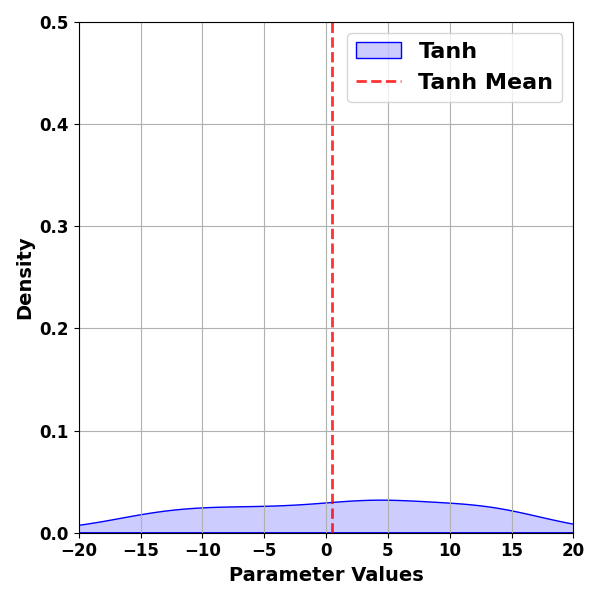}
      \end{subfigure}
    \end{minipage}
  \end{minipage}
\end{figure}

\subsubsection{PAU}
\begin{figure}[H]
\begin{minipage}[t]{01\textwidth}
  \vspace{0pt} 
    \begin{minipage}[t]{\textwidth}
      \centering
      \begin{subfigure}[t]{0.16\textwidth}
        \includegraphics[width=\linewidth]{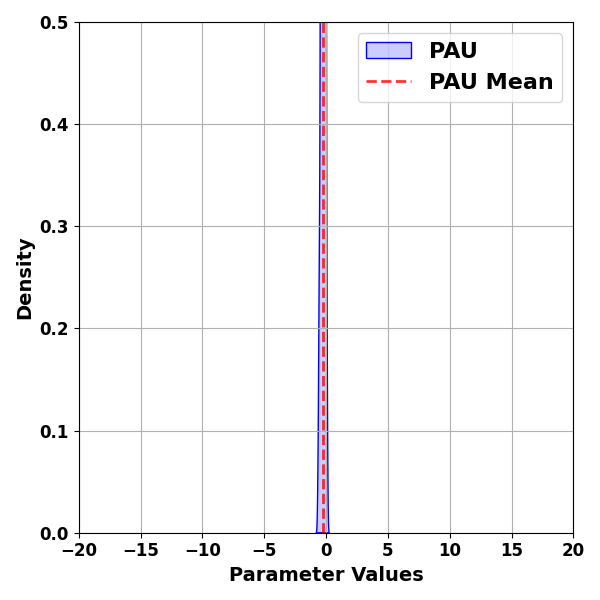}
      \end{subfigure}
      \hfill
     \begin{subfigure}[t]{0.16\textwidth}
        \includegraphics[width=\linewidth]{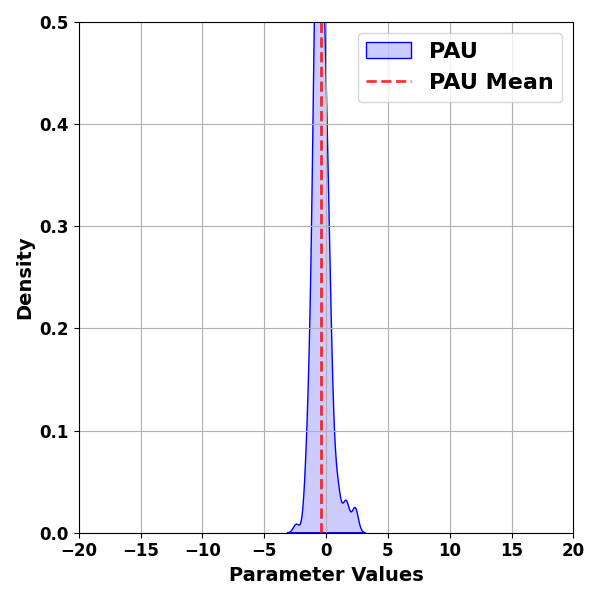}
      \end{subfigure}
      \hfill
      \begin{subfigure}[t]{0.16\textwidth}
        \includegraphics[width=\linewidth]{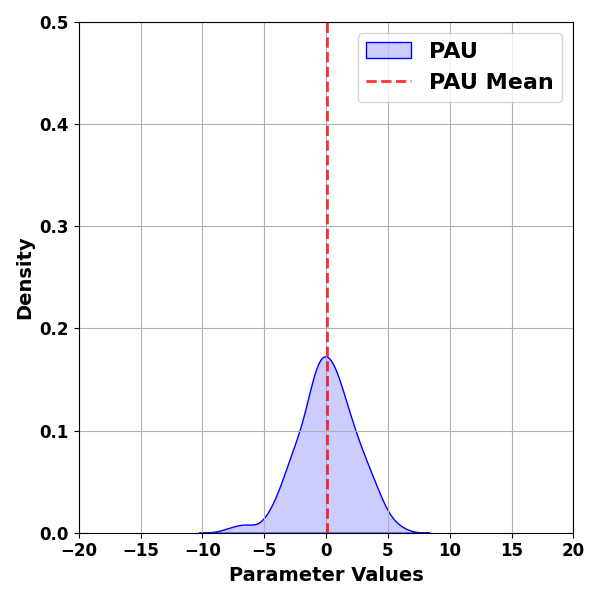}
      \end{subfigure}
      \hfill
      \begin{subfigure}[t]{0.16\textwidth}
        \includegraphics[width=\linewidth]{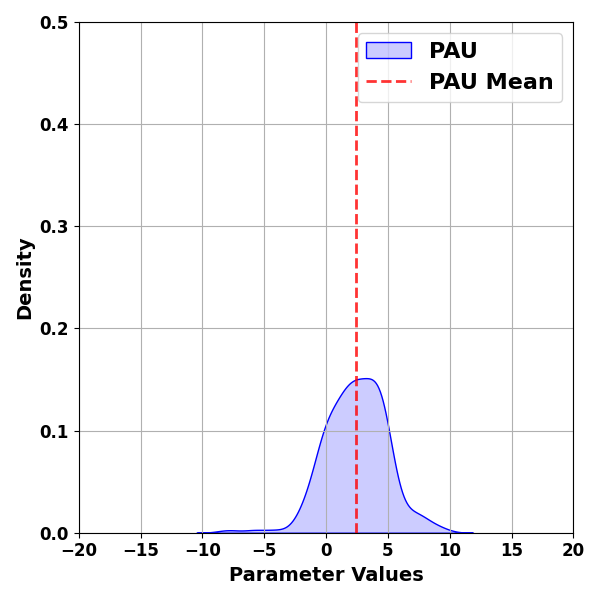}
      \end{subfigure}
      \hfill
      \begin{subfigure}[t]{0.16\textwidth}
        \includegraphics[width=\linewidth]{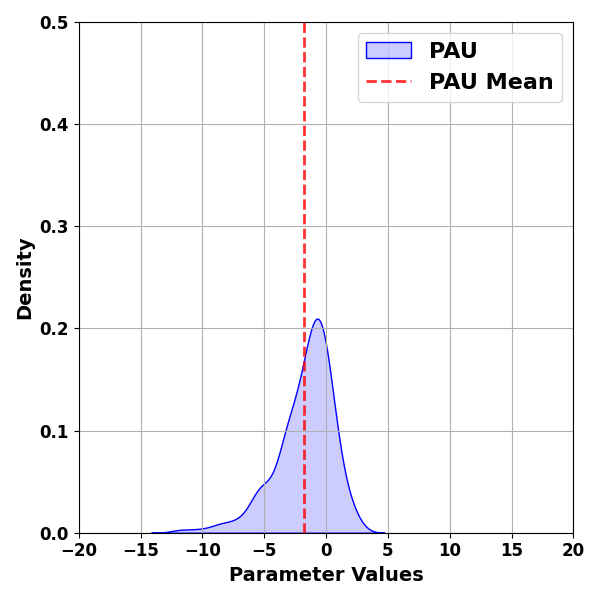}
      \end{subfigure}
      \hfill
      \begin{subfigure}[t]{0.16\textwidth}
        \includegraphics[width=\linewidth]{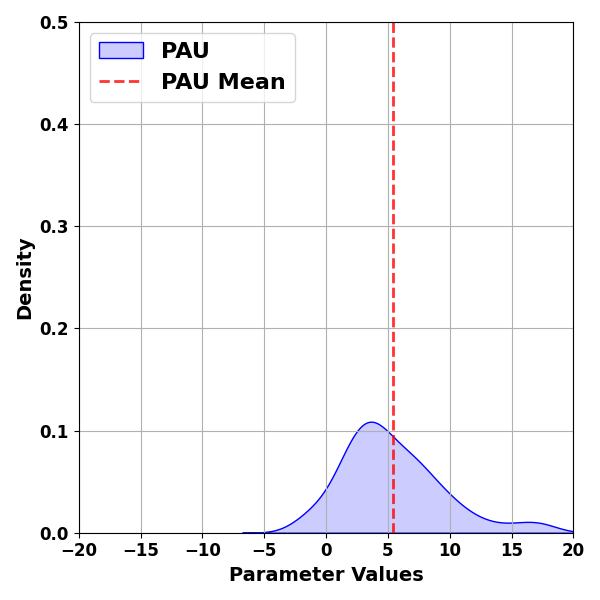}
      \end{subfigure}
    \end{minipage}
  \end{minipage}
\end{figure}

\subsubsection{ReLUDown}
\begin{figure}[H]
\begin{minipage}[t]{01\textwidth}
  \vspace{0pt} 
    \begin{minipage}[t]{\textwidth}
      \centering
      \begin{subfigure}[t]{0.16\textwidth}
        \includegraphics[width=\linewidth]{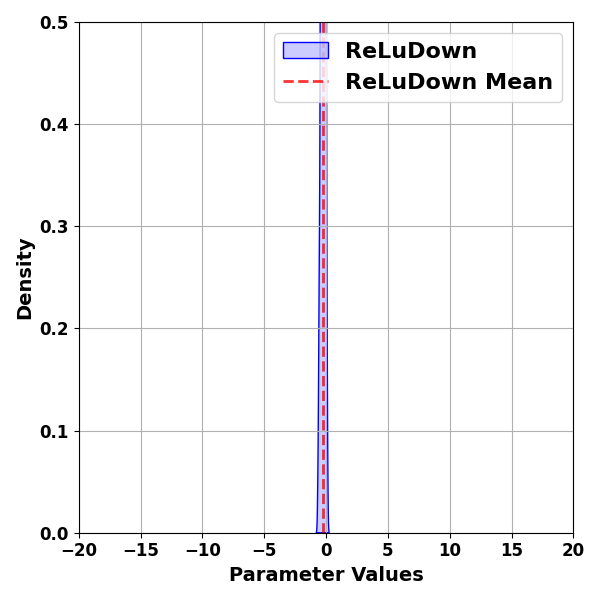}
      \end{subfigure}
      \hfill
     \begin{subfigure}[t]{0.16\textwidth}
        \includegraphics[width=\linewidth]{figures/activations/reludown/frame_100.png}
      \end{subfigure}
      \hfill
      \begin{subfigure}[t]{0.16\textwidth}
        \includegraphics[width=\linewidth]{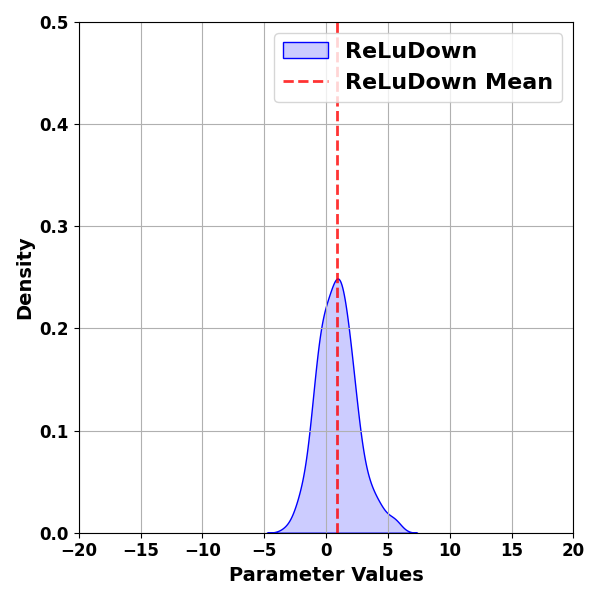}
      \end{subfigure}
      \hfill
      \begin{subfigure}[t]{0.16\textwidth}
        \includegraphics[width=\linewidth]{figures/activations/reludown/frame_1000.png}
      \end{subfigure}
      \hfill
      \begin{subfigure}[t]{0.16\textwidth}
        \includegraphics[width=\linewidth]{figures/activations/reludown/frame_3000.png}
      \end{subfigure}
      \hfill
      \begin{subfigure}[t]{0.16\textwidth}
        \includegraphics[width=\linewidth]{figures/activations/reludown/frame_5000.png}
      \end{subfigure}
    \end{minipage}
  \end{minipage}
\end{figure}
\subsection{Network and Hyperparameters}
\begin{table}[h!]
\centering
\begin{tabular}{|l|c|c|}
\hline
\multicolumn{3}{|l|}{\textbf{Network Layers}} \\
\hline
\textbf{Layer} & \textbf{Output Shape} & \textbf{Activation} \\
\hline
Input & $3 \times 32 \times 32$ & -- \\
Conv2d (3, 32, 5) & $32 \times 28 \times 28$ & ReLU, Tanh, PAU, ReLUDown\\
MaxPool2d (2, 2) & $32 \times 14 \times 14$ & -- \\
Conv2d (32, 64, 3) & $64 \times 12 \times 12$ & ReLU, Tanh, PAU, ReLUDown \\
MaxPool2d (2, 2) & $64 \times 6 \times 6$ & -- \\
Conv2d (64, 128, 3) & $128 \times 4 \times 4$ & ReLU, Tanh, PAU, ReLUDown \\
MaxPool2d (2, 2) & $128 \times 2 \times 2$ & -- \\
Flatten & $128 \times 4 = 512$ & -- \\
Linear (512, 128) & 128 & ReLU, Tanh, PAU, ReLUDown \\
Linear (128, 128) & 128 & ReLU, Tanh, PAU, ReLUDown \\
Linear (128, 2) & 2 & ReLU, Tanh, PAU, ReLUDown \\
\hline
\hline
\multicolumn{3}{|l|}{\textbf{Loss Function: Cross Entropy Loss}} \\
\hline
\textbf{Parameter} & \textbf{Symbol} & \textbf{Value} \\
\hline
Step Size & - & 0.010 \\
Weight Decay & - & 0.000 \\
Momentum & - & 0.900 \\
\hline
\hline
\multicolumn{3}{|l|}{\textbf{ReLuDown}} \\
\hline
\textbf{Parameter} & \textbf{Symbol} & \textbf{Value} \\
\hline
Hinge Point & $d$ & 3 \\
\hline
\hline
\multicolumn{3}{|l|}{\textbf{Decrease Backpropagation}} \\
\hline
\textbf{Parameter} & \textbf{Symbol} & \textbf{Value} \\
\hline
Decrease Factor & $f$ & 0.150 \\
Speed factor & $a$ & 1.005 \\
\hline
\hline
\multicolumn{3}{|l|}{\textbf{Padé Activation Units}} \\
\hline
\textbf{Parameter} & \textbf{Symbol} & \textbf{Value} \\
\hline
Nominator polynomial & $m$ & 5 \\
Denominator polynomial & $n$ & 4 \\
Initialization Shape   & -& ReLU \\
\hline
\hline
\multicolumn{3}{|l|}{\textbf{Continual Backpropagation}} \\
\hline
\textbf{Parameter} & \textbf{Symbol} & \textbf{Value} \\
\hline
Replacement Rate & - & 0.0001 \\
Decay Rate & - & 0.99 \\
Maturity Treshold   &- & 100 \\
Initialization &- & Kaiming\\
Utilization Type & -& Contribution\\
\hline
\hline
\multicolumn{3}{|l|}{\textbf{Generative Replay: Variational Autoencoder}} \\
\hline
\textbf{Parameter} & \textbf{Symbol} & \textbf{Value} \\
\hline
Step Size & - & 0.01 \\
Replay Percentage & - & 1/6 \\
\hline
\textbf{Layer} & \textbf{Output Shape} & \textbf{Activation} \\
\hline
Input (Image + Label) & $4 \times 32 \times 32$ & -- \\
Conv2d (4, 32, 4, 2, 1) & $32 \times 16 \times 16$ & ReLU \\
Conv2d (32, 64, 4, 2, 1) & $64 \times 8 \times 8$ & ReLU \\
Conv2d (64, 128, 4, 2, 1) & $128 \times 4 \times 4$ & ReLU \\
Flatten & 2048 & -- \\
FC → $\mu$, $\log\sigma^2$ & $2 \times 128$ & -- \\
\hline
$z$ + Label Embedding & 130 & -- \\
FC → 2048 & 2048 & -- \\
Unflatten & $128 \times 4 \times 4$ & -- \\
ConvT (128 → 64) & $64 \times 8 \times 8$ & ReLU \\
ConvT (64 → 32) & $32 \times 16 \times 16$ & ReLU \\
ConvT (32 → 3) & $3 \times 32 \times 32$ & Sigmoid \\
\hline
\end{tabular}
\end{table}

\newpage

\end{document}